%% file: preprint.tex
\documentclass[letterpaper]{article} 
\usepackage[preprint]{aaai2027}  
\usepackage{times}  
\usepackage{helvet}  
\usepackage{courier}  
\usepackage[hyphens]{url}  
\usepackage{graphicx} 
\usepackage{natbib}  
\usepackage{caption} 
\usepackage{algorithm}

\usepackage{amsmath}
\usepackage{amssymb}
\usepackage{amsfonts}
\usepackage{bm}
\usepackage{booktabs}
\usepackage{multirow}
\usepackage{amsthm}
\usepackage{algorithm}
\usepackage{algpseudocode}
\usepackage{microtype} 

\usepackage{amsmath}
\usepackage{booktabs}
\usepackage{enumitem}

\usepackage{algorithm}
\usepackage{algpseudocode}
\usepackage{pifont}        

\usepackage{graphicx}
\usepackage{amsmath}
\usepackage{multirow}
\usepackage{linguex}
\usepackage{sidecap}
\usepackage[utf8]{inputenc}
\usepackage[T1]{fontenc}

\usepackage{verbatim}

\usepackage{amssymb}
\usepackage{amsthm}

\usepackage{amsthm, amsmath, amssymb}
\newtheorem{proposition}{Proposition}

\usepackage[utf8]{inputenc} 
\usepackage[T1]{fontenc}    
\usepackage{booktabs}       
\usepackage{amsfonts}       
\usepackage{nicefrac}       
\usepackage{microtype}      
\usepackage{xcolor}         

\usepackage[most]{tcolorbox}
\tcbuselibrary{breakable}
\definecolor{promptbg}{RGB}{245, 248, 250}
\definecolor{promptborder}{RGB}{76, 175, 147}
\definecolor{prompttitle}{RGB}{76, 175, 147}

\usepackage{amsmath,amssymb,amsthm,bm}

\usepackage{verbatim}

\usepackage{booktabs}
\usepackage{multirow}
\usepackage{makecell}
\usepackage{xcolor}

\usepackage{amsmath}
\usepackage{amssymb}
\usepackage{amsthm}
\usepackage{xspace}
\usepackage{algorithm}
\usepackage{algpseudocode}

\newcommand{\method}{\textsc{CPSD}\xspace}
\newcommand{\sg}{\operatorname{sg}}

\usepackage{newfloat}
\usepackage{listings}
\DeclareCaptionStyle{ruled}{labelfont=normalfont,labelsep=colon,strut=off} 
\floatstyle{ruled}
\newfloat{listing}{tb}{lst}{}
\floatname{listing}{Listing}
\title{Contrastive On-Policy Self-Distillation for LLM Unlearning}

\author{
    Fangzhou Lin\textsuperscript{\rm 1,2,3,4},
    Shuo Xing\textsuperscript{\rm 1},
    Xin Li\textsuperscript{\rm 4}\thanks{Advising.},
    Wenbin He\textsuperscript{\rm 4},
    Zhengzhong Tu\textsuperscript{\rm 1}\footnotemark[1],
    Liu Ren\textsuperscript{\rm 4}\thanks{Corresponding authors.}
}
\affiliations{
    \textsuperscript{\rm 1}Texas A\&M University
    \textsuperscript{\rm 2}Worcester Polytechnic Institute
    \textsuperscript{\rm 3}Tohoku University\\
    \textsuperscript{\rm 4}Bosch Research North America \& Bosch Center for AI
}

\begin{document}

\maketitle

\input{sec/abstract}    

\input{sec/intro}

\input{sec/relatedwork}
\input{sec/method}
\input{sec/experiment}
\input{sec/conclusion}

\clearpage
\bibliography{aaai2026}
\clearpage
\appendix

\input{sec/appendix}

\end{document}

%% file: sec/abstract.tex
\begin{abstract}
Large language model unlearning aims to remove the influence of designated training data while preserving unrelated model capabilities. Existing likelihood-based methods commonly optimize against fixed forget sequences, providing only negative supervision on what the model should not generate and overlooking the states visited by the evolving unlearned policy. We introduce Contrastive On-Policy Self-Distillation (CPSD), a new unlearning framework that constructs an explicit knowledge-reduced target distribution and distills it along the model’s own trajectories. Given a student rollout, a frozen reference model evaluates each token under a plain context and a privileged context containing the forget information. CPSD treats the resulting distributional shift as a knowledge-associated direction and reverses it to obtain a counterfactual unlearning target. The student is then trained to match this target at every rollout position while retaining its utility on unrelated data. We further introduce a privacy-enhanced variant that combines distributional unlearning with lightweight sequence-level suppression. Experiments under OpenUnlearning show that CPSD improves the forgetting–utility trade-off over NPO across TOFU forget-set sizes and Llama models ranging from 1B to 8B parameters. On MUSE-News, the privacy-enhanced variant improves all reported knowledge, memorization, privacy, and utility metrics over NPO. \textbf{All code and materials will be open-sourced.}
\end{abstract}

%% file: sec/intro.tex
\section{Introduction}
\label{sec:introduction}

\begin{figure}[h]
    \centering
    \includegraphics[width=0.95\columnwidth]{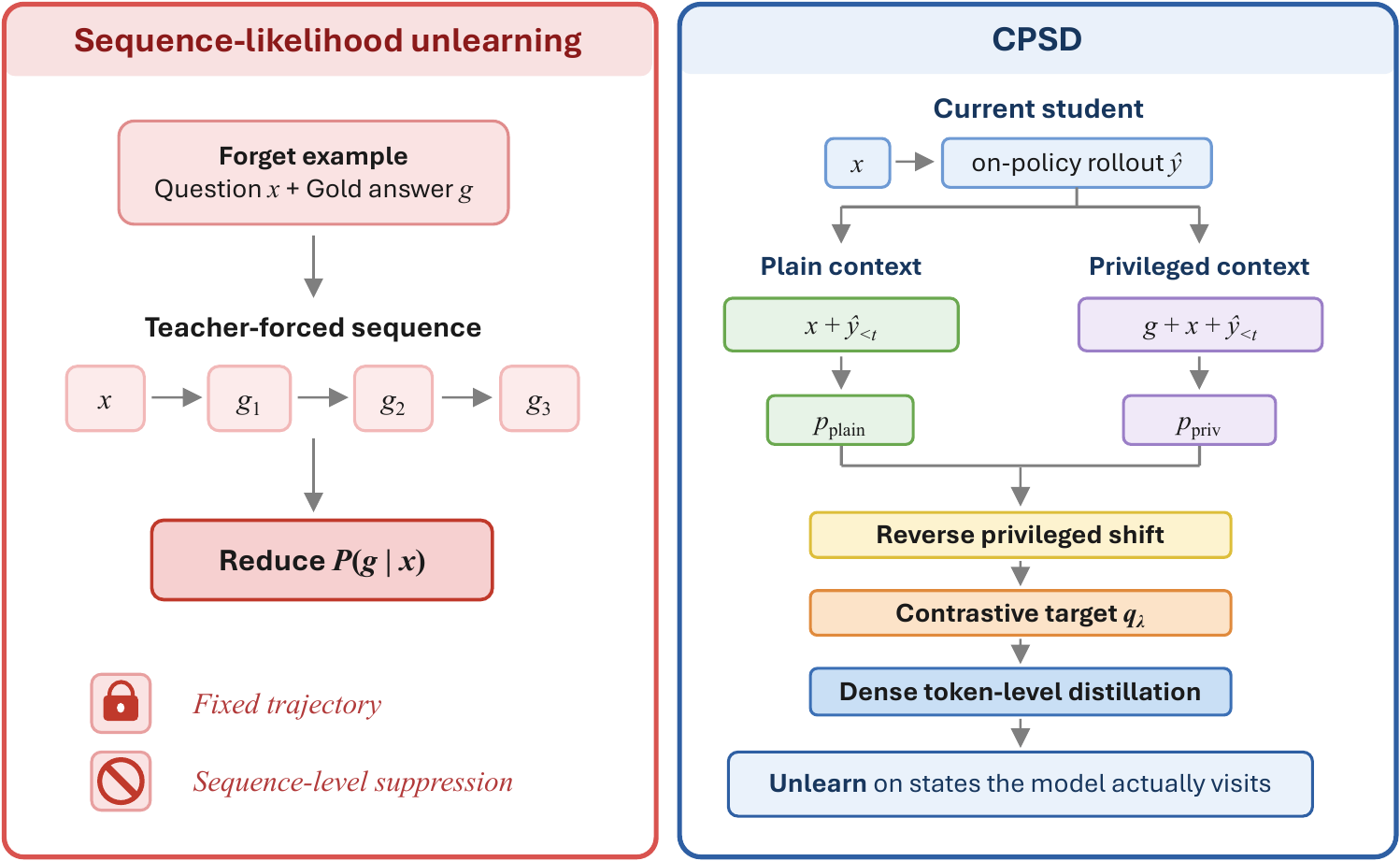}
    \caption{\textbf{Conceptual comparison between sequence-likelihood unlearning and CPSD}. Sequence-likelihood methods suppress dataset-provided forget
tokens along fixed, teacher-forced trajectories. CPSD instead samples
trajectories from the current student, contrasts the predictions of a
frozen reference model with and without privileged forget information,
and distills the reversed distributional shift along the student's own
states.
}
    \label{fig:target}
\vspace{-0.2cm}    
\end{figure}

Large language models (LLMs) acquire broad capabilities from massive training corpora, but may also memorize private information, copyrighted text, or other content that should later be removed\cite{naveed2025comprehensive,liu2025rethinking,zhang2025right,wang2025unique}.
Such memorization raises growing concerns regarding data privacy, intellectual property, model safety, and compliance with data-deletion requests.
Retraining an LLM from scratch whenever a subset of its training data must be deleted is prohibitively expensive, motivating the development of \emph{machine unlearning}\cite{bourtoule2021machine,shi2025muse,nguyen2025survey}.
The goal of LLM unlearning is to efficiently remove the influence of a designated forget set while preserving the model's knowledge and utility on unrelated data
\citep{cao2024fundamental, maini2024tofu, dorna2026openunlearning,liu2025rethinking}.

Existing weight-based LLM unlearning methods employ a variety of signals to guide parameter updates, including output likelihoods,
preference objectives, and intermediate representations\cite{ji2024reversing,jia2024wagle,li2025llm,fan2026simplicity}.
Sequence-likelihood methods such as Gradient Ascent directly decrease
the likelihood of forget examples, while preference-based objectives
such as Negative Preference Optimization (NPO)\cite{zhang2024negative} suppress their
likelihood relative to a reference model
\citep{yao2023editing,yao2024large,zhang2024negative,zhao2026unlearning}. Other approaches introduce alternative targets, such as refusal
responses, or define unlearning objectives over internal model
representations\cite{wang2023machine,shi2024ulmr,mekala2025alternate}.
For example, Representation Misdirection for Unlearning (RMU)
redirects forget-set representations at selected layers while encouraging representations on retain data to remain close to those of the original model
\citep{dang2025effects}.

Despite their effectiveness, widely used sequence-likelihood
objectives such as Gradient Ascent and NPO are evaluated primarily on
dataset-provided forget sequences under teacher forcing.
They provide a direct signal for reducing the probability of the
observed forget tokens, but do not explicitly construct a complete
next-token target distribution describing how the model should behave
after forgetting.
This distinction is important for autoregressive language models:
during deployment, the model generates its own response token by
token and therefore visits states induced by its previous predictions.
As the model changes during unlearning, the trajectories it generates
may also move away from the fixed sequences used by the unlearning
objective.
Consequently, optimizing forgetting only on dataset-provided
trajectories may leave the evolving model insufficiently guided at
the states it actually encounters.
A desirable unlearning method should therefore adapt its supervision
to model-generated trajectories while providing richer guidance than
decreasing the likelihood of individual forget tokens.

On-policy knowledge distillation offers a useful perspective for addressing this problem.
Unlike conventional distillation on fixed sequences, on-policy distillation allows the student model to generate its own trajectories and receive dense token-level supervision from a teacher along those trajectories
\citep{agarwal2024policy,li2026rethinking,fu2026revisiting}.
Recent work on On-Policy Self-Distillation (OPSD) further shows that the student and teacher need not be separate models:
the same LLM can act as a stronger teacher when it is given privileged information, such as a verified solution, while the student observes only the original question
\citep{zhao2026self}.
The privileged teacher then evaluates the student's own generation and guides it toward better reasoning.

Inspired by this learning paradigm, we ask:
\emph{Can an LLM also use its own knowledge to teach itself how to forget?}
A direct application of self-distillation is unsuitable for unlearning, because imitating a teacher that has access to the target information would reinforce that information.
Our key insight is to use the privileged teacher \emph{contrastively} rather than imitatively.
Given the same student-generated trajectory, we evaluate a frozen reference model under two conditions:
one with only the original input, and another with additional access to the forget information.
The change in the model's predictions reveals how the privileged information influences its output distribution.
Instead of teaching the student to follow this change, we reverse it to construct a target that moves the student away from the behavior induced by the forget information.

Based on this insight, we introduce
\emph{Contrastive On-Policy Self-Distillation} (\textsc{CPSD}), a new framework for LLM unlearning.
For each forget-set input, the current student model first generates an on-policy response.
A frozen copy of the original model then evaluates this response both with and without privileged forget information.
\textsc{CPSD} contrasts the two resulting predictions to derive a token-level unlearning target, and trains the student to follow this target throughout its own generated trajectory.
In this way, the model does not merely learn which fixed answer to suppress.
Instead, it receives dense supervision describing how its behavior should change at the states it actually visits.

The proposed design has three appealing properties.
First, it provides full-distribution token-level guidance rather than supervision only on the observed forget answer.
Second, it remains on-policy: as the student changes, the trajectories used for unlearning change with it.
Third, it requires no external teacher model, since both contrastive predictions are produced by the same frozen reference model under different contexts.
We jointly optimize a retain-set language-modeling objective to preserve unrelated capabilities.
For settings that explicitly evaluate exact-sequence memorization and membership privacy, we additionally introduce a privacy-enhanced variant that combines the on-policy distributional objective with lightweight sequence-level suppression.

\textsc{CPSD} can be applied to different forms of forget data.
For question-answering tasks, the answer associated with the forget example serves as privileged information.
For raw-text unlearning, an observed continuation from the forget corpus provides the privileged context, while the preceding text is used to elicit the student's response.
This allows the same framework to address both fine-grained knowledge unlearning and memorization-oriented removal from natural text.

We evaluate \textsc{CPSD} using the unified OpenUnlearning framework
\citep{dorna2026openunlearning}
on two complementary benchmarks.
TOFU evaluates fine-grained knowledge unlearning through question answering
\citep{maini2024tofu},
while MUSE evaluates knowledge, verbatim memorization, privacy, and utility on raw text
\citep{shi2025muse}.
On TOFU, \textsc{CPSD} consistently improves the forgetting--utility trade-off over NPO across multiple forget-set sizes and Llama models ranging from 1B to 8B parameters.
In particular, on the larger \texttt{forget10} setting, \textsc{CPSD} achieves a better Forget Quality--Model Utility operating point than NPO at every evaluated model scale.
On MUSE-News, the privacy-enhanced variant improves all reported forgetting, retention, extraction, and privacy metrics over the corresponding NPO baseline.
These results demonstrate that contrastive on-policy self-distillation provides an effective and scalable alternative to fixed-sequence likelihood suppression.

Our main contributions are summarized as follows:
\begin{itemize}
    \item
    We formulate LLM unlearning as \emph{contrastive on-policy self-distillation}, enabling the model to learn how to forget along trajectories generated by its evolving policy rather than only on fixed forget sequences.

    \item
    We introduce a contrastive self-teacher that evaluates the same student trajectory with and without privileged forget information.
    By reversing the behavioral shift induced by privileged information, \textsc{CPSD} constructs dense token-level guidance for unlearning without requiring an external teacher model.

    \item
    We develop a unified framework for both question-answering and raw-text unlearning, together with a privacy-enhanced extension for exact-sequence removal.
    Experiments on TOFU and MUSE demonstrate consistent improvements over NPO across different forget-set sizes, model scales, and evaluation dimensions.
\end{itemize}

%% file: sec/relatedwork.tex
\section{Related Work}
\label{sec:related_work}

\paragraph{LLM unlearning.}
Existing LLM unlearning methods mainly optimize output likelihoods,
preferences, or internal representations\cite{si2023knowledge,jia2024soul,liu2024machine,geng2025comprehensive,li2025machine,liu2025rethinking,dorna2026openunlearning}.
Gradient Ascent and GradDiff reduce the likelihood of forget examples,
while NPO and SimNPO use bounded preference-style objectives
\citep{maini2024tofu,liu2022continual,fan2026simplicity,zhao2026unlearning}.
Other methods construct alternative output targets or modify hidden
representations, such as UNDIAL and RMU
\citep{dong2025undial, li2024wmdp}.
Unlike these approaches, \method constructs a full-vocabulary
unlearning target on trajectories sampled from the evolving student.

\paragraph{On-policy self-distillation.}
On-policy distillation trains a student on its own generated
trajectories, reducing the mismatch between training and inference
states \citep{czarnecki2019distilling,agarwal2024policy,ding2026hdpo,song2026survey,ye2026policy,zhu2026many,yang2026learning,yang2025survey,yang2026self}.
OPSD further uses privileged information to let a frozen copy of the
initial model supervise the student without an external teacher
\citep{zhao2026self}.
\method adopts this on-policy self-distillation framework, but uses the
privileged prediction contrastively: instead of imitating it, we reverse
the shift between privileged and plain reference predictions to construct
an unlearning target.

%% file: sec/method.tex
%

\section{Contrastive On-Policy Self-Distillation}
\label{sec:method}


\subsection{Problem Setup and Method Overview}
\label{sec:method_overview}

Let $\pi_{\theta_0}$ denote the model before unlearning.
We are given a forget set
$\mathcal D_f=\{(x,g)\}$ and a retain set
$\mathcal D_r=\{(x^r,y^r)\}$.
Here, $x$ is an eliciting context and $g$ is the associated content
whose influence should be removed.
For question-answering data, $x$ is a question prompt and $g$ is its
answer; for raw text, $x$ is a text prefix and $g$ is an observed
continuation from the forget corpus.

We initialize a trainable student $\pi_\theta$ and a frozen reference
model $\pi_\phi$ from the same pre-unlearning parameters:
$\theta\leftarrow\theta_0$ and $\phi\leftarrow\theta_0$.
The student is updated throughout unlearning, whereas the reference
model remains fixed and provides a stable description of the original
model behavior.

Our method, \emph{Contrastive On-Policy Self-Distillation}
(\method), consists of three stages.
First, the current student generates a response from the forget prompt,
ensuring that unlearning is performed at states visited by the evolving
student policy.
Second, the frozen reference evaluates the same student trajectory
under two contexts: a plain context containing only the original prompt,
and a privileged context that additionally contains the forget content
$g$.
Third, \method contrasts the resulting predictions and reverses the
distributional change induced by the privileged information.
The student is then distilled toward this contrastive target along its
own generated trajectory.

This design differs from standard self-distillation\cite{zhang2019your,mobahi2020self,zhang2021self}.
A conventional privileged teacher is directly imitated because its
additional information improves the desired prediction.
For unlearning, directly imitating such a teacher would instead reinforce
the content to be removed.
\method therefore uses privileged information only to identify how the
original model's output distribution changes when the target content is
available, and constructs the learning target in the opposite direction.

\subsection{Contrastive Target on Student Trajectories}
\label{sec:contrastive_target}

For each forget example $(x,g)$, the current student samples an
autoregressive trajectory
\begin{equation}
\label{eq:rollout}
    \hat{\mathbf y}
    =
    (\hat y_1,\ldots,\hat y_m)
    \sim
    \pi_{\sg(\theta)}(\cdot\mid x),
\end{equation}
where $\sg$ indicates that gradients do not propagate through the
discrete generation process.
The sampled trajectory is treated as fixed during the subsequent
optimization step.
At position $t$, the student distribution is
\begin{equation}
\label{eq:student_distribution}
    s_{\theta,t}(v)
    =
    \pi_\theta
    \left(
        v\mid x,\hat{\mathbf y}_{<t}
    \right),
    \qquad v\in\mathcal V.
\end{equation}

The frozen reference model evaluates the same rollout prefix under two
conditions:
\begin{align}
\label{eq:reference_distributions}
    p_t^{0}(v)
    &=
    \pi_{\phi,T}
    \left(
        v\mid x,\hat{\mathbf y}_{<t}
    \right), \\
    p_t^{+}(v)
    &=
    \pi_{\phi,T}
    \left(
        v\mid c(g),x,\hat{\mathbf y}_{<t}
    \right),
\end{align}
where $c(g)$ is a fixed template containing the privileged forget
content, and $T$ is the reference temperature.
Importantly, $p_t^0$ and $p_t^+$ are aligned to exactly the same
student-generated position, so that their difference reflects the
effect of adding the privileged context rather than a difference in
generated trajectories.

We define the privileged-context influence of token $v$ as
\begin{equation}
\label{eq:influence}
    \rho_t(v)
    =
    \log p_t^{+}(v)
    -
    \log p_t^{0}(v).
\end{equation}
A positive $\rho_t(v)$ indicates that access to $g$ increases the
reference model's relative support for token $v$.
We treat $\rho_t$ as a knowledge-associated output-distribution signal,
without assuming that it represents a unique internal knowledge
direction.

Given a contrast strength $\lambda\geq0$, \method constructs the
normalized target
\begin{equation}
\label{eq:cpsd_target}
    q_{\lambda,t}(v)
    =
    \frac{
        p_t^{0}(v)
        \exp\!\left[-\lambda\rho_t(v)\right]
    }{
        \sum_{u\in\mathcal V}
        p_t^{0}(u)
        \exp\!\left[-\lambda\rho_t(u)\right]
    }.
\end{equation}
Equivalently,
\begin{equation}
\label{eq:cpsd_target_ratio}
    q_{\lambda,t}(v)
    \propto
    \frac{
        p_t^{0}(v)^{1+\lambda}
    }{
        p_t^{+}(v)^\lambda
    }.
\end{equation}

Equation~\eqref{eq:cpsd_target} can be viewed as an exponential tilting
of the plain reference distribution.
The plain prediction $p_t^0$ serves as a behavioral anchor, while tokens
that are relatively amplified by the privileged information are
downweighted.
When $\lambda=0$, $q_{\lambda,t}$ reduces to $p_t^0$.
Increasing $\lambda$ more strongly reverses the relative preference
induced by the privileged context.


\begin{proposition}[Minimal contrastive intervention]
\label{prop:variational_target}
For a fixed rollout state, the target in
Eq.~\eqref{eq:cpsd_target} is the unique solution to
\begin{equation}
\label{eq:variational_target}
    \arg\min_{q\in\Delta(\mathcal V)}
    \left\{
        D_{\mathrm{KL}}(q\|p_t^0)
        +
        \lambda
        \mathbb E_{v\sim q}
        [\rho_t(v)]
    \right\}.
\end{equation}
\end{proposition}

Thus, \method selects the distribution that remains close to the plain
reference while assigning less probability to tokens strongly supported
by the privileged forget information.
The proof and further properties of $\lambda$ are given in
Appendix ~\ref{app:target_properties}.

\subsection{Unlearning and Retention Objectives}
\label{sec:training_objective}

The student is distilled toward the contrastive target at every rollout
position:
\begin{align}
\label{eq:cpsd_loss}
    \mathcal L_{\mathrm{CPSD}}
    =
    \mathbb E_{\substack{
        (x,g)\sim\mathcal D_f\\
        \hat{\mathbf y}\sim
        \pi_{\sg(\theta)}(\cdot\mid x)
    }}
    \left[
        \frac{1}{|\hat{\mathbf y}|}
        \sum_{t=1}^{|\hat{\mathbf y}|}
        D_{\mathrm{KL}}
        \left(
            s_{\theta,t}
            \middle\|
            q_{\lambda,t}
        \right)
    \right].
\end{align}
We explicitly use
$D_{\mathrm{KL}}(s_{\theta,t}\|q_{\lambda,t})$\cite{perez2008kullback}.
Gradients flow only through the student distribution; the rollout,
reference predictions, and contrastive target are stop-gradient
quantities within each optimization step.

Substituting Eq.~\eqref{eq:cpsd_target} yields
\begin{align}
\label{eq:loss_decomposition}
    D_{\mathrm{KL}}
    \left(
        s_{\theta,t}\|q_{\lambda,t}
    \right)
    &=
    D_{\mathrm{KL}}
    \left(
        s_{\theta,t}\|p_t^0
    \right)
    +
    \lambda
    \mathbb E_{v\sim s_{\theta,t}}
    [\rho_t(v)]
    +
    \log Z_t,
\end{align}
where $Z_t$ is the normalization term in
Eq.~\eqref{eq:cpsd_target} and is independent of $\theta$.
Therefore, minimizing the CPSD objective can be interpreted as
simultaneously anchoring the student to the plain reference and reducing
its expected support for tokens favored by privileged forget
information.
Unlike a scalar sequence-level signal, $\rho_t(v)$ provides dense
full-vocabulary guidance at every student-generated position.

To preserve unrelated capabilities, we jointly optimize the standard
language-modeling objective\cite{guu2020retrieval}
\begin{equation}
\label{eq:retain_loss}
    \mathcal L_{\mathrm{ret}}
    =
    -
    \mathbb E_{(x^r,y^r)\sim\mathcal D_r}
    \left[
        \frac{1}{|y^r|}
        \sum_t
        \log
        \pi_\theta
        \left(
            y_t^r\mid x^r,y_{<t}^r
        \right)
    \right].
\end{equation}
The basic training objective is
\begin{equation}
\label{eq:total_loss}
    \mathcal L
    =
    \gamma\mathcal L_{\mathrm{CPSD}}
    +
    \alpha\mathcal L_{\mathrm{ret}}.
\end{equation}

For settings that explicitly evaluate exact-sequence memorization or
membership privacy, we additionally use a lightweight NPO regularizer\citep{zhang2024negative}:
\begin{equation}
\label{eq:hybrid_loss}
    \mathcal L_{\mathrm{CPSD\text{-}H}}
    =
    \gamma
    \left(
        \mathcal L_{\mathrm{CPSD}}
        +
        \eta\mathcal L_{\mathrm{NPO}}
    \right)
    +
    \alpha\mathcal L_{\mathrm{ret}}.
\end{equation}
The two objectives act at complementary granularities:
$\mathcal L_{\mathrm{CPSD}}$ reshapes the full output distribution on
student-generated trajectories, whereas
$\mathcal L_{\mathrm{NPO}}$ directly suppresses the likelihood of the
dataset-provided forget sequence.
The complete NPO formulation is deferred to
Appendix~\ref{app:privacy_objective}.


Algorithm~\ref{alg:cpsd} summarizes the training procedure.
The same formulation applies to both question-answering and raw-text
unlearning.
For question answering, $g$ is the target answer; for raw text, $g$ is
an observed continuation following the eliciting prefix $x$.
Detailed implementations  are provided in
Appendix~\ref{app:implementation}.

\begin{algorithm}[t]
\caption{Contrastive On-Policy Self-Distillation}
\label{alg:cpsd}
\begin{algorithmic}[1]
\Require Forget set $\mathcal D_f$, retain set $\mathcal D_r$,
student $\pi_\theta$, frozen reference $\pi_\phi$
\While{not converged}
    \State Sample forget batch $\mathcal B_f$ and retain batch
    $\mathcal B_r$
    \State Generate
    $\hat{\mathbf y}\sim
    \pi_{\sg(\theta)}(\cdot\mid x)$
    for $(x,g)\in\mathcal B_f$
    \State Evaluate $s_{\theta,t}$, $p_t^0$, and $p_t^+$
    on the same rollout positions
    \State Compute
    $\rho_t=\log p_t^+-\log p_t^0$
    \State Construct
    $q_{\lambda,t}\propto p_t^0\exp(-\lambda\rho_t)$
    \State Compute $\mathcal L_{\mathrm{CPSD}}$ and
    $\mathcal L_{\mathrm{ret}}$
    \State Optionally add $\eta\mathcal L_{\mathrm{NPO}}$
    \State Update $\theta$ and keep $\phi$ frozen
\EndWhile
\end{algorithmic}
\end{algorithm}

%% file: sec/experiment.tex
\section{Experiments}
\label{sec:experiments}


\subsection{Experimental Setup}
\label{sec:experimental_setup}

\paragraph{Benchmarks.}
We conduct experiments on two complementary benchmarks implemented in
the OpenUnlearning framework \citep{dorna2026openunlearning}.
\textbf{TOFU} \citep{maini2024tofu} evaluates fine-grained unlearning
of fictitious author profiles represented as question-answer pairs.
We consider both \texttt{forget05} and \texttt{forget10}, which require
removing $5\%$ and $10\%$ of the injected author knowledge,
respectively.
\textbf{MUSE} \citep{shi2025muse} evaluates unlearning from natural
text with separate measurements of verbatim memorization, semantic
knowledge, privacy, and retained utility.
We use the MUSE-News split to test whether CPSD extends beyond
question-answering data to pretraining-style text.

\paragraph{Models.}
On TOFU, we evaluate Llama-3.2-1B, Llama-3.2-3B, and Llama-3.1-8B\cite{grattafiori2024llama},
using the corresponding full-data task models released through
OpenUnlearning.
This range allows us to examine whether the behavior of CPSD is
consistent from small to substantially larger LLMs.
On MUSE-News, we follow the benchmark setup and use Llama-2-7B\cite{touvron2023llama}.

\paragraph{Baselines.}
Our primary comparison is Negative Preference Optimization (NPO)
\citep{zhang2024negative}, a strong likelihood-based unlearning method that
suppresses forget sequences relative to a frozen reference model.
All methods are implemented and evaluated within the same
OpenUnlearning pipeline.
The final comparison additionally includes representative
likelihood-, preference-, distillation-, and representation-based
methods, including Gradient Ascent(GA)\cite{maini2024tofu}, GradDiff\cite{liu2022continual}, SimNPO\cite{fan2026simplicity}, UNDIAL\cite{dong2025undial}, and RMU\cite{li2024wmdp}.
%

\paragraph{Evaluation metrics.}
On TOFU, we report \emph{Forget Quality} (FQ) and
\emph{Model Utility} (MU)\cite{maini2024tofu}.
FQ is based on a statistical comparison between the truth-ratio
distributions of the unlearned model and a retain-only model; higher
values indicate that the two distributions are more similar.
MU aggregates performance on retained authors, real-world authors, and
general factual questions, with higher values indicating better
capability preservation.
For privacy-oriented analysis, we additionally report
probability(Prob.)\cite{maini2024tofu} and ROUGE\cite{lin2004rouge}, for which lower is better, and Privacy Leakage
(PrivLeak)\cite{shi2025muse}, for which larger values indicate greater similarity to the
retain-only model in the TOFU implementation.

On MUSE, lower forget-set verbatim memorization
(\textsc{fVerb}\cite{dorna2026openunlearning}), knowledge memorization (\textsc{fKnow})\cite{shi2025muse}, and
extraction strength (\textsc{Extr})\cite{carlini2021extracting} indicate stronger forgetting. For MUSE PrivLeak, values closer to zero are better.
We explicitly distinguish these conventions because the reported
PrivLeak scales differ between TOFU and MUSE.

\paragraph{Implementation details.}
Unless stated otherwise, we use contrast strength $\lambda=8$,
teacher temperature $T=1$, and equal forget and retain weights
$\gamma=\alpha=1$.
The student-to-target divergence is
$D_{\mathrm{KL}}(s_{\theta,t}\|q_{\lambda,t})$, and no divergence
clipping is applied.
Student rollouts contain at most 128 new tokens and are sampled with
temperature $1.0$, top-$p$ $0.95$, and top-$k$ $50$.
We select the learning rate from
$\{5\times10^{-5}, 1\times10^{-4}\}$.
The selected TOFU configurations use $1\times10^{-4}$ for the 1B and
3B models and either $5\times10^{-5}$ or $1\times10^{-4}$ for the 8B
model depending on the split.
MUSE experiments use $1\times10^{-4}$.
For the privacy-enhanced variant, we use NPO weight $\eta=0.3$ on the
TOFU privacy experiment and $\eta=0.1$ on MUSE-News.
Additional optimization and data-processing details are provided in
Appendix~\ref{app:implementation}.

\subsection{Main Results on TOFU}
\label{sec:tofu_main}

Table~\ref{tab:tofu_main} presents the main TOFU results across two
forget-set sizes and three model scales.
On the more challenging \texttt{forget10} setting, CPSD obtains a
strictly better numbers than NPO and other six methods for every evaluated
model on both FQ and MU.
For Llama-3.2-1B, CPSD increases FQ from
$9.5\times10^{-5}$ to $3.13\times10^{-3}$ while improving MU from
$0.391$ to $0.517$.
The improvement is particularly pronounced for Llama-3.2-3B, where FQ
increases from $1.65\times10^{-8}$ to $0.155$ and MU simultaneously
increases from $0.552$ to $0.615$.
The benefit persists at 8B parameters: CPSD improves both FQ and MU,
showing that its effectiveness is not restricted to smaller models.





\begin{table}[h]
\centering
\small
\setlength{\tabcolsep}{.5pt}
\renewcommand{\arraystretch}{1.08}
\caption{
Main results on TOFU across model scales and forget-set sizes.
FQ and MU are both higher-is-better.
Bold values denote the results of our method, CPSD.
}
\label{tab:tofu_main}
\begin{tabular}{llcc@{\hspace{0.85cm}}cc}
\toprule
& &
\multicolumn{2}{c}{\texttt{forget05}} &
\multicolumn{2}{c}{\texttt{forget10}} \\
\cmidrule(lr){3-4}
\cmidrule(lr){5-6}
Model & Method &
FQ $\uparrow$ & MU $\uparrow$ &
FQ $\uparrow$ & MU $\uparrow$ \\
\midrule

\multirow{7}{*}{Lla-3.2-1B}
& GA
& $0.012$ & $0.402$
& $1.20{\times}10^{-5}$ & $0.356$ \\
& GradDiff
& $0.086$ & $0.468$
& $4.80{\times}10^{-4}$ & $0.472$ \\
& NPO
& $4.75{\times}10^{-5}$ & $0.450$
& $9.50{\times}10^{-5}$ & $0.391$ \\
& SimNPO
& $0.174$ & $0.486$
& $1.65{\times}10^{-3}$ & $0.501$ \\
& UNDIAL
& $0.109$ & $0.474$
& $7.60{\times}10^{-4}$ & $0.483$ \\
& RMU
& $0.193$ & $0.491$
& $2.42{\times}10^{-3}$ & $0.510$ \\
& \textbf{CPSD}
& $\mathbf{0.221}$ & $\mathbf{0.497}$
& $\mathbf{3.13{\times}10^{-3}}$ & $\mathbf{0.517}$ \\
\midrule

\multirow{7}{*}{Lla-3.2-3B}
& GA
& $1.70{\times}10^{-4}$ & $0.501$
& $0.003$ & $0.506$ \\
& GradDiff
& $0.009$ & $0.551$
& $0.061$ & $0.579$ \\
& NPO
& $2.41{\times}10^{-8}$ & $0.547$
& $1.65{\times}10^{-8}$ & $0.552$ \\
& SimNPO
& $0.027$ & $0.571$
& $0.119$ & $0.606$ \\
& UNDIAL
& $0.016$ & $0.559$
& $0.084$ & $0.592$ \\
& RMU
& $0.034$ & $0.576$
& $0.143$ & $0.612$ \\
& \textbf{CPSD}
& $\mathbf{0.040}$ & $\mathbf{0.580}$
& $\mathbf{0.155}$ & $\mathbf{0.615}$ \\
\midrule

\multirow{7}{*}{Lla-3.1-8B}
& GA
& $0.041$ & $0.402$
& $2.00{\times}10^{-8}$ & $0.542$ \\
& GradDiff
& $0.266$ & $0.472$
& $3.80{\times}10^{-6}$ & $0.623$ \\
& NPO
& $6.49{\times}10^{-4}$ & $0.547$
& $1.80{\times}10^{-5}$ & $0.648$ \\
& SimNPO
& $0.518$ & $0.507$
& $2.70{\times}10^{-5}$ & $0.660$ \\
& UNDIAL
& $0.337$ & $0.486$
& $8.90{\times}10^{-6}$ & $0.638$ \\
& RMU
& $0.672$ & $0.512$
& $4.40{\times}10^{-5}$ & $0.666$ \\
& \textbf{CPSD}
& $\mathbf{0.713}$ & $\mathbf{0.564}$
& $\mathbf{5.00{\times}10^{-5}}$ & $\mathbf{0.669}$ \\
\bottomrule
\end{tabular}
\vspace{-0.3cm}
\end{table}

The same pattern holds on \texttt{forget05}.
CPSD substantially increases FQ at all three model scales.
At 1B and 3B parameters, these gains are accompanied by higher model
utility.
For the 8B model, CPSD raises FQ from
$6.49\times10^{-4}$ to $0.713$, while MU decreases moderately from
$0.547$ to $0.515$.
This point represents a more aggressive forgetting configuration rather
than a strict Pareto improvement, but the utility remains close to the
NPO operating point despite a large improvement in FQ.

Overall, the results provide two forms of consistency.
First, CPSD remains effective as the deletion request grows from
\texttt{forget05} to \texttt{forget10}.
Second, its advantage is observed across 1B, 3B, and 8B models rather
than being tied to a particular capacity.
This supports the central hypothesis that constructing an explicit
on-policy distributional target provides a more informative forgetting
signal than optimizing only on fixed forget sequences.


\subsection{Privacy-Oriented Unlearning}
\label{sec:tofu_privacy}

CPSD changes the model's full next-token distribution on self-generated
trajectories, while privacy metrics can additionally depend on the
likelihood of the exact training sequence.
We therefore evaluate the privacy-enhanced variant
\textsc{CPSD-H}, which adds a lightweight NPO regularizer to the
on-policy distributional objective.

Table~\ref{tab:tofu_privacy} reports a privacy-oriented operating point
on Llama-3.1-8B with \texttt{forget10}.
Compared with NPO, \textsc{CPSD-H} improves every reported
forgetting or privacy metric.
FQ increases from $1.8\times10^{-5}$ to $1.3\times10^{-3}$,
while forget probability and ROUGE decrease from $0.077$ and $0.485$
to $0.067$ and $0.240$, respectively.
Most notably, PrivLeak improves from $-26.0$ to $18.9$, moving much
closer to the retain-model regime.

\begin{table}[h]
\centering
\small

\setlength{\tabcolsep}{3.3pt}
\caption{
Privacy-oriented results on TOFU \texttt{forget10} with
Llama-3.1-8B.
For FQ, MU, and PrivLeak, higher is better; for probability and ROUGE,
lower is better.
}
\label{tab:tofu_privacy}
\begin{tabular}{lccccc}
\toprule
Method &
FQ $\uparrow$ &
Prob. $\downarrow$ &
ROUGE $\downarrow$ &
MU $\uparrow$ &
PrivLeak $\uparrow$ \\
\midrule
NPO
& $1.8{\times}10^{-5}$
& $0.077$
& $0.485$
& $\mathbf{0.648}$
& $-26.0$ \\
\textsc{CPSD-H}
& $\mathbf{1.3{\times}10^{-3}}$
& $\mathbf{0.067}$
& $\mathbf{0.240}$
& $0.480$
& $\mathbf{18.9}$ \\
\bottomrule
\end{tabular}
\end{table}

This stronger removal comes with a reduction in MU from $0.648$ to
$0.480$.
We therefore regard \textsc{CPSD-H} as a privacy-oriented operating
point rather than a universally dominant replacement for basic CPSD.
The result illustrates that the two objectives are complementary:
the CPSD component reshapes behavior on the model's own trajectories,
whereas the sequence-level component directly reduces the likelihood of
the original forget answers.


\begin{figure*}[h]
    \centering
    \includegraphics[width=0.85\textwidth]{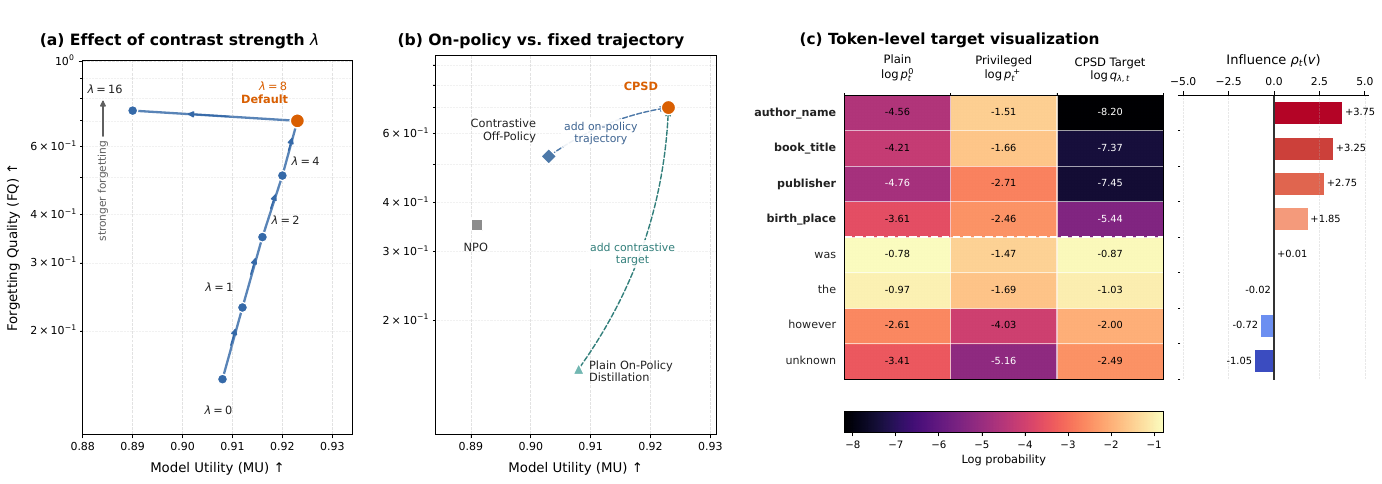}
    \caption{\textbf{Analysis of the CPSD mechanism.} (a) Forgetting--utility operating points as the contrast strength
$\lambda$ varies.
(b) Component analysis separating the effects of on-policy sampling and
contrastive target construction.
(c) Token-level plain, privileged, and CPSD target distributions for a
representative forget example. Tokens most strongly amplified by the
privileged context receive substantially lower probability under the
contrastive target.}
    \label{fig:analysis}
\vspace{-0.2cm}    
\end{figure*}

\subsection{Generalization to Raw-Text Unlearning}
\label{sec:muse_results}

We next evaluate whether CPSD depends on the question-answer structure
of TOFU.
For MUSE-News, we divide each forget-text chunk into an eliciting prefix
and a privileged continuation.
The student generates from the prefix, while the continuation is
provided only to the privileged reference.
All other components of the method remain unchanged.

Table~\ref{tab:muse_news} reports the results.
Basic CPSD improves all three direct removal metrics relative to NPO:
forget verbatim memorization decreases from $0.256$ to $0.236$,
forget knowledge memorization decreases from $0.361$ to $0.329$, and
extraction decreases from $0.057$ to $0.041$.
This confirms that the contrastive target remains effective when the
privileged information is a natural-text continuation rather than a
short answer.

\begin{table}[h]
\centering
\small
\setlength{\tabcolsep}{4.5pt}
\caption{
Results on MUSE-News with Llama-2-7B.
Lower \textsc{fVerb}, \textsc{fKnow}, and \textsc{Extr} are better; PrivLeak should be closer to zero.
}
\label{tab:muse_news}
\begin{tabular}{lccccc}
\toprule
Method &
\textsc{fVerb} $\downarrow$ &
\textsc{fKnow} $\downarrow$ &
PrivLeak $\rightarrow 0$ &
\textsc{Extr} $\downarrow$ \\
\midrule
NPO
& $0.256$
& $0.361$
& $-89.1$
& $0.057$ \\
\method
& $\mathbf{0.236}$
& $0.329$
& $-97.4$
& $\mathbf{0.041}$ \\
\textsc{CPSD-H}
& $0.242$
& $\mathbf{0.291}$
& $\mathbf{-88.9}$
& $0.047$ \\
\bottomrule
\end{tabular}
\end{table}

Basic CPSD trades some retained knowledge and privacy for stronger
direct removal.
Adding a small sequence-level weight, $\eta=0.1$, produces a more
balanced solution.
\textsc{CPSD-H} improves all five reported metrics over NPO:
it reduces both verbatim and knowledge memorization, moves PrivLeak closer to zero, and lowers
extraction strength.
The result demonstrates that the two-level objective generalizes across
both benchmark and data format.

Importantly, no task-specific refusal response or externally generated
negative answer is required.
The same privileged-contrast construction is instantiated using an
answer on TOFU and a continuation on MUSE.
This supports the interpretation of CPSD as a general distributional
unlearning framework rather than a QA-specific prompting strategy.

\subsection{Analysis}
\label{sec:experimental_analysis}

We complement the aggregate results with analyses designed to test the
mechanism underlying CPSD in Fig~\ref{fig:analysis}.


\paragraph{Effect of contrast strength.}
We vary $\lambda$ while keeping the optimization configuration fixed in Fig~\ref{fig:analysis} a.
The theoretical analysis in Appendix~\ref{app:lambda_properties}
predicts that increasing $\lambda$ more strongly suppresses expected
privileged influence at the target-distribution level.
The corresponding experiment measures whether this produces a
controllable empirical transition from preservation to forgetting.
We report FQ, MU, and, where applicable, PrivLeak as functions of
$\lambda$.

\paragraph{On-policy versus fixed trajectories.}
To isolate the contribution of on-policy sampling, we compare CPSD
against an otherwise identical variant that constructs and distills the
contrastive target on dataset-provided trajectories in Fig~\ref{fig:analysis} b.
A Pareto comparison of forgetting and utility determines whether
matching the states visited by the evolving student contributes beyond
the contrastive target itself.

\paragraph{Visualizing the contrastive target.}
For representative forget prompts, in Fig~\ref{fig:analysis} c, we visualize the top tokens under
the plain distribution $p_t^0$, privileged distribution $p_t^+$, and
contrastive target $q_{\lambda,t}$.
We additionally report the tokens with the largest privileged influence
$\rho_t(v)$.
The expected pattern is that tokens strongly amplified by access to the
forget content lose relative probability under the CPSD target, while
tokens unaffected by the privileged context remain anchored to the
plain distribution.


%% file: sec/conclusion.tex
\section{Conclusion}
\label{sec:conclusion}

We introduced \method, a contrastive on-policy self-distillation framework for LLM unlearning. By comparing a frozen reference model with and without privileged forget information, \method constructs a knowledge-reduced target distribution and distills it along trajectories generated by the evolving student. This formulation provides dense token-level guidance while preserving unrelated capabilities through a retain objective, and can be extended with sequence-level regularization for privacy-sensitive settings. Experiments on TOFU and MUSE demonstrate consistent improvements across model scales, forget-set sizes, and both question-answering and raw-text unlearning scenarios.

%% file: sec/appendix.tex

\section{Properties of the Contrastive Target}\label{app:target_properties}

This section provides additional derivations and formal properties of
the contrastive target introduced in
Eq.~\eqref{eq:cpsd_target}.
All results below characterize the target distribution at a fixed
student-generated state.
They do not imply exact equivalence to retraining on the retain set or
guarantee complete removal of the forget data.

\subsection{Equivalent Forms of the Target}
\label{app:target_equivalence}

At rollout position $t$, recall the plain and privileged reference
distributions
\begin{equation}
    p_t^0(v)
    =
    \pi_{\phi,T}
    \left(
        v\mid x,\hat{\mathbf y}_{<t}
    \right),
    \qquad
    p_t^+(v)
    =
    \pi_{\phi,T}
    \left(
        v\mid c(g),x,\hat{\mathbf y}_{<t}
    \right).
\end{equation}
The privileged-context influence score is
\begin{equation}
\label{eq:appendix_rho}
    \rho_t(v)
    =
    \log p_t^+(v)-\log p_t^0(v).
\end{equation}

The implementation first constructs the unnormalized log-score
\begin{equation}
\label{eq:appendix_target_score}
    a_{\lambda,t}(v)
    =
    (1+\lambda)\log p_t^0(v)
    -
    \lambda\log p_t^+(v),
\end{equation}
and then normalizes it over the vocabulary:
\begin{equation}
\label{eq:appendix_target_softmax}
    q_{\lambda,t}(v)
    =
    \frac{
        \exp(a_{\lambda,t}(v))
    }{
        \sum_{u\in\mathcal V}
        \exp(a_{\lambda,t}(u))
    }.
\end{equation}
Substituting Eq.~\eqref{eq:appendix_rho} into
Eq.~\eqref{eq:appendix_target_score} gives
\begin{align}
    a_{\lambda,t}(v)
    &=
    \log p_t^0(v)-\lambda\rho_t(v),
\end{align}
and therefore
\begin{equation}
\label{eq:appendix_exponential_tilt}
    q_{\lambda,t}(v)
    =
    \frac{
        p_t^0(v)\exp[-\lambda\rho_t(v)]
    }{
        Z_t(\lambda)
    },
\end{equation}
where
\begin{equation}
\label{eq:appendix_partition}
    Z_t(\lambda)
    =
    \sum_{u\in\mathcal V}
    p_t^0(u)\exp[-\lambda\rho_t(u)].
\end{equation}
Equivalently,
\begin{equation}
\label{eq:appendix_ratio_target}
    q_{\lambda,t}(v)
    =
    \frac{
        p_t^0(v)^{1+\lambda}
        p_t^+(v)^{-\lambda}
    }{
        \sum_{u\in\mathcal V}
        p_t^0(u)^{1+\lambda}
        p_t^+(u)^{-\lambda}
    }.
\end{equation}

The construction also admits an equivalent logit-space expression.
Let $z_t^0$ and $z_t^+$ denote the reference logits under the plain
and privileged contexts, respectively:
\begin{equation}
    p_t^0
    =
    \operatorname{softmax}(z_t^0/T),
    \qquad
    p_t^+
    =
    \operatorname{softmax}(z_t^+/T).
\end{equation}
Terms arising from the individual softmax normalization constants are
independent of the vocabulary item and disappear after the final
normalization. Hence,
\begin{equation}
\label{eq:appendix_logit_target}
    q_{\lambda,t}
    =
    \operatorname{softmax}
    \left(
        \frac{
            (1+\lambda)z_t^0-\lambda z_t^+
        }{T}
    \right).
\end{equation}
Equations~\eqref{eq:appendix_exponential_tilt} and
\eqref{eq:appendix_logit_target} are therefore mathematically
equivalent.

\subsection{Variational Characterization}
\label{app:variational_proof}

We now prove Proposition~\ref{prop:variational_target} from the main
text.

\begin{proof}
For readability, we omit the position subscript $t$.
Consider
\begin{equation}
\label{eq:appendix_variational_problem}
    \min_{q\in\Delta(\mathcal V)}
    \left\{
        D_{\mathrm{KL}}(q\|p^0)
        +
        \lambda\mathbb E_{v\sim q}[\rho(v)]
    \right\}.
\end{equation}
Introducing a Lagrange multiplier $\nu$ for the normalization
constraint gives
\begin{align}
    \mathcal J(q,\nu)
    &=
    \sum_{v\in\mathcal V}
    q(v)\log\frac{q(v)}{p^0(v)}
    +
    \lambda
    \sum_{v\in\mathcal V}
    q(v)\rho(v)
    \nonumber\\
    &\quad+
    \nu
    \left(
        \sum_{v\in\mathcal V}q(v)-1
    \right).
\end{align}
The first-order condition for each $v$ is
\begin{equation}
    \frac{\partial\mathcal J}{\partial q(v)}
    =
    \log\frac{q(v)}{p^0(v)}
    +
    1+\lambda\rho(v)+\nu
    =
    0.
\end{equation}
Rearranging yields
\begin{equation}
    q(v)
    =
    p^0(v)
    \exp[-\lambda\rho(v)]
    \exp[-1-\nu].
\end{equation}
The final factor is independent of $v$ and is determined by
normalization. Therefore,
\begin{equation}
    q(v)
    =
    \frac{
        p^0(v)\exp[-\lambda\rho(v)]
    }{
        \sum_{u\in\mathcal V}
        p^0(u)\exp[-\lambda\rho(u)]
    },
\end{equation}
which is exactly the CPSD target.

Since $D_{\mathrm{KL}}(q\|p^0)$ is strictly convex in $q$ when $p^0$
has full support, and the expectation term is linear in $q$, the
objective in Eq.~\eqref{eq:appendix_variational_problem} is strictly
convex. The solution is therefore unique.
\end{proof}

This result gives a precise interpretation of the target construction.
The KL term constrains the target to remain close to the original
plain-context behavior, whereas the second term penalizes probability
assigned to tokens whose support is increased by privileged forget
information.

\subsection{Pairwise-Odds Interpretation}
\label{app:pairwise_odds}

For any two tokens $u,v\in\mathcal V$, the normalization constant
cancels in their probability ratio:
\begin{align}
\label{eq:appendix_pairwise_odds}
    \log
    \frac{
        q_{\lambda,t}(v)
    }{
        q_{\lambda,t}(u)
    }
    &=
    \log
    \frac{
        p_t^0(v)
    }{
        p_t^0(u)
    }
    -
    \lambda
    \left[
        \rho_t(v)-\rho_t(u)
    \right].
\end{align}
Consequently, if
\begin{equation}
    \rho_t(v)>\rho_t(u),
\end{equation}
then
\begin{equation}
    \frac{\partial}{\partial\lambda}
    \log
    \frac{
        q_{\lambda,t}(v)
    }{
        q_{\lambda,t}(u)
    }
    <0.
\end{equation}
Thus, increasing $\lambda$ monotonically decreases the target odds of
tokens that are more strongly amplified by the privileged context,
relative to tokens that are less strongly amplified.

This property concerns the constructed target distribution. It does
not imply that an external unlearning metric must vary monotonically
with $\lambda$, since the final trained model also depends on
optimization, retention regularization, and model capacity.

\subsection{Monotonic Control of Privileged Influence}
\label{app:lambda_properties}

The contrast strength $\lambda$ has a stronger distribution-level
interpretation than the pairwise result above.
Define the expected privileged influence under the target as
\begin{equation}
    \mu_t(\lambda)
    =
    \mathbb E_{v\sim q_{\lambda,t}}
    [\rho_t(v)].
\end{equation}

\begin{proposition}[Monotonic privileged-influence reduction]
\label{prop:lambda_monotonicity}
For every fixed rollout state,
\begin{equation}
\label{eq:appendix_mu_derivative}
    \frac{d\mu_t(\lambda)}{d\lambda}
    =
    -
    \operatorname{Var}_{v\sim q_{\lambda,t}}
    [\rho_t(v)]
    \leq 0.
\end{equation}
Moreover, for
\begin{equation}
    K_t(\lambda)
    =
    D_{\mathrm{KL}}
    \left(
        q_{\lambda,t}\|p_t^0
    \right),
\end{equation}
we have, for $\lambda\geq0$,
\begin{equation}
\label{eq:appendix_kl_derivative}
    \frac{dK_t(\lambda)}{d\lambda}
    =
    \lambda
    \operatorname{Var}_{v\sim q_{\lambda,t}}
    [\rho_t(v)]
    \geq0.
\end{equation}
\end{proposition}

\begin{proof}
Again omitting $t$, define
\begin{equation}
    Z(\lambda)
    =
    \sum_v
    p^0(v)\exp[-\lambda\rho(v)].
\end{equation}
Its log derivative is
\begin{align}
    \frac{d}{d\lambda}\log Z(\lambda)
    &=
    \frac{
        -\sum_v
        p^0(v)\rho(v)\exp[-\lambda\rho(v)]
    }{
        Z(\lambda)
    }
    \nonumber\\
    &=
    -
    \mathbb E_{v\sim q_\lambda}[\rho(v)]
    =
    -\mu(\lambda).
\end{align}
Differentiating once more gives
\begin{equation}
    \frac{d^2}{d\lambda^2}
    \log Z(\lambda)
    =
    \operatorname{Var}_{q_\lambda}[\rho].
\end{equation}
It follows that
\begin{equation}
    \mu'(\lambda)
    =
    -
    \operatorname{Var}_{q_\lambda}[\rho],
\end{equation}
which proves Eq.~\eqref{eq:appendix_mu_derivative}.

Using
\begin{equation}
    \log
    \frac{q_\lambda(v)}{p^0(v)}
    =
    -\lambda\rho(v)-\log Z(\lambda),
\end{equation}
we obtain
\begin{equation}
    K(\lambda)
    =
    -\lambda\mu(\lambda)-\log Z(\lambda).
\end{equation}
Therefore,
\begin{align}
    K'(\lambda)
    &=
    -\mu(\lambda)
    -
    \lambda\mu'(\lambda)
    -
    \frac{d}{d\lambda}\log Z(\lambda)
    \nonumber\\
    &=
    -\lambda\mu'(\lambda)
    =
    \lambda
    \operatorname{Var}_{q_\lambda}[\rho],
\end{align}
which proves Eq.~\eqref{eq:appendix_kl_derivative}.
\end{proof}

Proposition~\ref{prop:lambda_monotonicity} formalizes the role of
$\lambda$ at the target-distribution level.
Increasing $\lambda$ reduces the expected privileged-context influence,
while increasing the deviation from the plain reference.
It therefore directly controls a local contrast between influence
removal and preservation of the plain-context distribution.

\subsection{KL-Regularized Dense-Signal Interpretation}
\label{app:dense_signal}

For any student distribution $s_{\theta,t}$, the CPSD divergence can be
expanded using
Eq.~\eqref{eq:appendix_exponential_tilt}:
\begin{align}
    D_{\mathrm{KL}}
    \left(
        s_{\theta,t}\|q_{\lambda,t}
    \right)
    &=
    \sum_v
    s_{\theta,t}(v)
    \log
    \frac{
        s_{\theta,t}(v)
    }{
        q_{\lambda,t}(v)
    }
    \nonumber\\
    &=
    D_{\mathrm{KL}}
    \left(
        s_{\theta,t}\|p_t^0
    \right)
    +
    \lambda
    \mathbb E_{v\sim s_{\theta,t}}
    [\rho_t(v)]
    \nonumber\\
    &\quad+
    \log Z_t(\lambda).
\label{eq:appendix_loss_decomposition}
\end{align}
Because $p_t^0$, $\rho_t$, and $Z_t$ are produced by the frozen
reference model, the last term is constant with respect to the student
parameters at a fixed rollout state. Therefore,
\begin{align}
    \arg\min_{s}
    D_{\mathrm{KL}}(s\|q_{\lambda,t})
    =
    \arg\max_s
    \left\{
        -\lambda\mathbb E_{v\sim s}[\rho_t(v)]
        -
        D_{\mathrm{KL}}(s\|p_t^0)
    \right\}.
\label{eq:appendix_regularized_reward}
\end{align}
Defining a vocabulary-level signal
\begin{equation}
    R_t(v)
    =
    -\lambda\rho_t(v),
\end{equation}
CPSD can be interpreted as maximizing a dense negative-privilege
signal while regularizing the student toward the plain reference
distribution.

Equation~\eqref{eq:appendix_regularized_reward} is a per-state
distributional identity.
CPSD samples states from the current student but stops gradients through
the discrete rollout.
We therefore do not claim that the implemented objective computes the
full policy gradient of a trajectory-level return.

\section{Privacy-Enhanced Objective}
\label{app:privacy_objective}

The basic CPSD objective reshapes the full next-token distribution on
student-generated trajectories.
For evaluations sensitive to the likelihood of the original forget
sequence, we augment it with a bounded NPO term.

For a forget example $(x,g)$, define the average conditional
negative log-likelihood under the student as
\begin{equation}
\label{eq:appendix_student_nll}
    \ell_\theta(x,g)
    =
    -
    \frac{1}{|g|}
    \sum_{j=1}^{|g|}
    \log
    \pi_\theta
    \left(
        g_j\mid x,g_{<j}
    \right).
\end{equation}
The corresponding frozen-reference NLL is
\begin{equation}
\label{eq:appendix_reference_nll}
    \ell_\phi(x,g)
    =
    -
    \frac{1}{|g|}
    \sum_{j=1}^{|g|}
    \log
    \pi_\phi
    \left(
        g_j\mid x,g_{<j}
    \right).
\end{equation}
Let
\begin{equation}
    \delta_\theta(x,g)
    =
    \ell_\theta(x,g)-\ell_\phi(x,g).
\end{equation}
The sequence-level regularizer is
\begin{equation}
\label{eq:appendix_npo}
    \mathcal L_{\mathrm{NPO}}
    =
    -
    \frac{2}{\beta_{\mathrm N}}
    \mathbb E_{(x,g)\sim\mathcal D_f}
    \left[
        \log\sigma
        \left(
            \beta_{\mathrm N}
            \delta_\theta(x,g)
        \right)
    \right],
\end{equation}
where $\beta_{\mathrm N}>0$ controls the saturation rate.

The sign of Eq.~\eqref{eq:appendix_npo} can be verified directly:
\begin{equation}
    \frac{\partial\mathcal L_{\mathrm{NPO}}}
         {\partial\delta_\theta}
    =
    -2
    \sigma
    \left(
        -\beta_{\mathrm N}\delta_\theta
    \right)
    <0.
\end{equation}
Gradient descent therefore increases $\delta_\theta$, encouraging
\begin{equation}
    \ell_\theta(x,g)>\ell_\phi(x,g),
\end{equation}
which decreases the likelihood of the forget sequence relative to the
pre-unlearning reference model.

The complete privacy-enhanced objective is
\begin{equation}
\label{eq:appendix_hybrid_objective}
    \mathcal L_{\mathrm{CPSD-H}}
    =
    \gamma
    \left(
        \mathcal L_{\mathrm{CPSD}}
        +
        \eta\mathcal L_{\mathrm{NPO}}
    \right)
    +
    \alpha\mathcal L_{\mathrm{ret}},
\end{equation}
where $\eta$ controls the strength of exact-sequence suppression.
Setting $\eta=0$ recovers the basic CPSD objective.

The two forget losses use different training states.
$\mathcal L_{\mathrm{CPSD}}$ is evaluated on student-generated
trajectories and supplies a full-vocabulary signal, whereas
$\mathcal L_{\mathrm{NPO}}$ is evaluated under teacher forcing on the
dataset-provided forget sequence.
Their combination therefore integrates on-policy behavioral reshaping
with direct sequence-level likelihood suppression.

\section{Implementation Details}
\label{app:implementation}

\subsection{Extracting Eliciting and Privileged Content}
\label{app:data_construction}

CPSD requires each forget example to provide an eliciting context $x$
and associated privileged content $g$.
The extraction procedure depends on the format of the benchmark.

\paragraph{Question-answering data.}
For tokenized question-answering examples, answer tokens are identified
by positions whose language-model labels are not equal to $-100$.
If $j$ is the first answer position, we define
\begin{equation}
    x
    =
    \texttt{input\_ids}_{<j},
    \qquad
    g
    =
    \texttt{input\_ids}_{\{\texttt{labels}\neq-100\}}.
\end{equation}
Thus, the student generates from the question prompt alone, while the
answer is used only as privileged content and, for CPSD-H, as the
sequence-level forget target.

\paragraph{Raw-text data.}
Raw pretraining-style examples do not contain a separate prompt and
answer.
Given the non-padding tokens
\begin{equation}
    \mathbf w=(w_1,\ldots,w_L),
\end{equation}
we set
\begin{equation}
\label{eq:appendix_text_k}
    k
    =
    \min
    \left(
        K_{\max},
        \max
        \left(
            1,
            \left\lfloor\frac{L}{2}\right\rfloor
        \right)
    \right),
\end{equation}
and define
\begin{equation}
\label{eq:appendix_text_split}
    x
    =
    w_{1:k},
    \qquad
    g
    =
    w_{k+1:\min(L,k+H_{\max})}.
\end{equation}
In our MUSE experiments, we use
$K_{\max}=512$ and $H_{\max}=256$.
The student generates a new continuation from $x$, whereas the observed
continuation $g$ is provided only to the privileged reference and the
optional sequence-level objective.

\subsection{Privileged-Context Construction}
\label{app:privileged_prompt}

The privileged content is inserted into a fixed template
\begin{equation}
    c(g)
    =
    c_{\mathrm{pre}}
    \oplus g
    \oplus c_{\mathrm{suf}},
\end{equation}
where $\oplus$ denotes token concatenation.
The default prefix indicates that the following span contains relevant
reference information, and the suffix separates that information from
the original prompt.

For a minibatch, privileged blocks may have different lengths.
Each block is left-padded to a common width $B$ and then prepended to
the student sequence:
\begin{equation}
    \mathbf z^{+}
    =
    [c(g)\oplus x\oplus\hat{\mathbf y}].
\end{equation}
The same frozen model $\pi_\phi$ is used for both plain and privileged
evaluation.
No gradients are propagated through either reference forward pass.

\subsection{Rollout Generation}
\label{app:rollout_details}

Prompts are left-padded to a common length $P$ before batched
generation.
The student samples with
\begin{equation}
    \hat{\mathbf y}
    \sim
    \pi_{\sg(\theta)}(\cdot\mid x),
\end{equation}
using stochastic decoding.
Our default generation configuration uses a maximum of 128 new tokens,
temperature $1.0$, nucleus threshold $0.95$, and top-$k$ value $50$.

Generation is performed without gradients and with key--value caching
enabled.
After generation, the model is returned to its original training mode,
and the sampled sequence is scored again in a differentiable student
forward pass.

\subsection{Position IDs under Left Padding}
\label{app:position_ids}

Left padding is required for batched autoregressive generation.
To ensure that non-padding tokens receive consistent positions, we
construct position IDs from the attention mask:
\begin{equation}
\label{eq:appendix_position_ids}
    \operatorname{pos}_{b,j}
    =
    \max
    \left(
        0,
        \sum_{r\leq j}
        \operatorname{mask}_{b,r}
        -1
    \right).
\end{equation}
Padding positions are assigned position zero.
These position IDs are supplied to the student, plain-reference, and
privileged-reference forward passes.

\subsection{Logit Alignment}
\label{app:logit_alignment}

Let $P$ be the padded prompt width and let the complete student sequence
be
\begin{equation}
    \mathbf z
    =
    [x\oplus\hat{\mathbf y}].
\end{equation}
The logits predicting the rollout tokens are selected as
\begin{equation}
\label{eq:appendix_plain_slice}
    \mathbf l_{\mathrm{student}},
    \mathbf l_{\mathrm{plain}}
    =
    \operatorname{logits}(\mathbf z)
    [:,P-1:-1,:].
\end{equation}
The shift by one position ensures that each selected vector predicts
the corresponding rollout token.

The privileged input contains an additional padded block of width $B$.
Its aligned rollout logits are therefore
\begin{equation}
\label{eq:appendix_priv_slice}
    \mathbf l_{\mathrm{priv}}
    =
    \operatorname{logits}(\mathbf z^+)
    [:,B+P-1:-1,:].
\end{equation}
Equations~\eqref{eq:appendix_plain_slice} and
\eqref{eq:appendix_priv_slice} select predictions for exactly the same
rollout positions.
This alignment is essential: comparing distributions at different
autoregressive states would confound privileged-context influence with
trajectory differences.

Only valid generated tokens contribute to the CPSD objective.
Padding positions in the completion are assigned label $-100$ and
excluded from the token average.

\subsection{Temperature and Divergence}
\label{app:divergence_details}

The two reference distributions are computed as
\begin{equation}
    \log p_t^0
    =
    \operatorname{logsoftmax}
    \left(
        z_t^0/T
    \right),
    \qquad
    \log p_t^+
    =
    \operatorname{logsoftmax}
    \left(
        z_t^+/T
    \right).
\end{equation}
The target score is then
\begin{equation}
    a_{\lambda,t}
    =
    (1+\lambda)\log p_t^0
    -
    \lambda\log p_t^+.
\end{equation}
Because temperature scaling has already been applied to the reference
distributions, the target is passed to the distribution-matching loss
with temperature $1$ to avoid applying the temperature twice.

The implementation supports a generalized divergence family, but all
main CPSD experiments use
\begin{equation}
    D_{\mathrm{KL}}
    \left(
        s_{\theta,t}\|q_{\lambda,t}
    \right).
\end{equation}
We write the two arguments explicitly rather than using the terms
``forward KL'' or ``reverse KL,'' whose conventions differ across the
distillation literature.

The final CPSD configuration does not clip the per-position divergence.
The loss is averaged over all valid rollout positions:
\begin{equation}
    \mathcal L_{\mathrm{CPSD}}
    =
    \frac{
        \sum_{b,t}
        m_{b,t}
        D_{\mathrm{KL}}
        \left(
            s_{\theta,b,t}\|q_{\lambda,b,t}
        \right)
    }{
        \sum_{b,t}m_{b,t}
    },
\end{equation}
where $m_{b,t}\in\{0,1\}$ is the completion mask.

\subsection{Detailed Optimization Procedure}
\label{app:detailed_algorithm}

\begin{algorithm}[t]
\caption{Detailed CPSD Training Step}
\label{alg:cpsd_detailed}
\begin{algorithmic}[1]
\Require Forget batch $\mathcal B_f$, retain batch $\mathcal B_r$,
student $\pi_\theta$, frozen reference $\pi_\phi$
\State Extract $(x,g)$ from each forget example
\State Left-pad prompts and construct prompt attention masks
\State Sample student rollouts
$\hat{\mathbf y}\sim\pi_{\sg(\theta)}(\cdot\mid x)$
\State Construct completion masks for valid rollout tokens
\State Score $[x\oplus\hat{\mathbf y}]$ with $\pi_\theta$
to obtain $s_{\theta,t}$
\State Score $[x\oplus\hat{\mathbf y}]$ with $\pi_\phi$
to obtain $p_t^0$
\State Construct privileged blocks
$c(g)=c_{\mathrm{pre}}\oplus g\oplus c_{\mathrm{suf}}$
\State Score
$[c(g)\oplus x\oplus\hat{\mathbf y}]$
with $\pi_\phi$ to obtain $p_t^+$
\State Align all three distributions to the same rollout positions
\State Compute
$\rho_t=\log p_t^+-\log p_t^0$
\State Construct
$q_{\lambda,t}\propto p_t^0\exp(-\lambda\rho_t)$
\State Compute
$\mathcal L_{\mathrm{CPSD}}$
over valid rollout positions
\If{$\eta>0$}
    \State Compute $\mathcal L_{\mathrm{NPO}}$
    on the original forget sequence
\EndIf
\State Compute retain NLL
$\mathcal L_{\mathrm{ret}}$ on $\mathcal B_r$
\State Form the total objective and update only $\theta$
\end{algorithmic}
\end{algorithm}

\subsection{Gradient Flow}
\label{app:gradient_flow}

Within each training step, the following quantities are treated as
stop-gradient:
\begin{equation}
    \hat{\mathbf y},
    \qquad
    p_t^0,
    \qquad
    p_t^+,
    \qquad
    \rho_t,
    \qquad
    q_{\lambda,t}.
\end{equation}
Gradients from the CPSD loss propagate only through
$s_{\theta,t}$.
Gradients from the retain loss propagate through the student evaluation
on the retain batch.
When CPSD-H is enabled, gradients from the NPO term propagate through
the student likelihood of the forget sequence, while the corresponding
reference likelihood remains fixed.

The reference parameters $\phi$ are never updated:
\begin{equation}
    \nabla_\phi\mathcal L=0.
\end{equation}
This prevents the contrastive target from drifting together with the
unlearned student.

\subsection{Computational Cost}
\label{app:computational_cost}

A basic CPSD training step requires:

\begin{enumerate}
    \item one autoregressive student generation on the forget prompts;
    \item one differentiable student forward pass on the sampled
    trajectories;
    \item two non-differentiable reference forward passes, corresponding
    to the plain and privileged contexts; and
    \item one differentiable student forward pass on the retain batch.
\end{enumerate}

Let $C_{\mathrm{gen}}$ denote the cost of rollout generation,
$C_{\mathrm{roll}}$ the cost of scoring a rollout sequence, and
$C_{\mathrm{ret}}$ the cost of the retain forward pass.
The high-level per-step computation is
\begin{equation}
    O
    \left(
        C_{\mathrm{gen}}
        +
        3C_{\mathrm{roll}}
        +
        C_{\mathrm{ret}}
    \right).
\end{equation}
The two reference forwards require no activation storage for
backpropagation.
In practice, autoregressive rollout generation is the principal
additional cost relative to teacher-forced unlearning methods.

CPSD-H additionally evaluates the student and frozen reference on the
dataset-provided forget sequence in order to compute the NPO
regularizer.